\documentclass[journal]{IEEEtran}
\usepackage{caption}
\usepackage{tikz}
\usepackage{booktabs}
\usepackage{amsmath}
\usepackage[linesnumbered,ruled]{algorithm2e}
\usepackage{algpseudocode}
\usepackage{tabulary}
     
\usepackage{booktabs}   
\usepackage{multirow}   
\usepackage{caption} 
\usepackage{mathtools}

\usepackage{array,tabularx}

\usepackage{cite}

\usepackage{graphicx}
\usepackage{amssymb}
\usepackage{floatrow}
\usepackage{amsmath}
\usepackage{lipsum}
\usepackage{multicol}
\usepackage{tikz}
\newcommand{\nonl}{\renewcommand{\nl}{\let\nl\oldnl}}

\usetikzlibrary{arrows,shapes,positioning,shadows,trees}

\tikzset{
  basic/.style  = {draw, text width=2cm, drop shadow, font=\scriptsize, rectangle},
  root/.style   = {basic, rounded corners=2pt, thin, align=center,
                   fill=white!30},
  level 2/.style = {basic, rounded corners=4pt, thin,align=center, fill=white!60,
                   text width=5em},
  level 3/.style = {basic, thin, align=left, fill=white!60, text width=5em}
}

\newcommand{\CASE}[1]{\STATE \textbf{case} #1\textbf{:} \begin{ALC@g}}
\newcommand{\ENDCASE}{\end{ALC@g}}

\newcommand{\DEFAULT}{\STATE \textbf{default:} \begin{ALC@g}}
\newcommand{\ENDDEFAULT}{\end{ALC@g}}
\newcommand{\DEFAULTLINE}[1]{\STATE \textbf{default:} }

\ifCLASSINFOpdf
\else
\fi
\begin{document}
%

\title{A Multi-objective Optimization Approach for Feature Selection in Gentelligent Systems}

%
%

\author{Mohammadhossein Ghahramani,~\IEEEmembership{Senior Member,~IEEE,} Yan Qiao,~\IEEEmembership{Senior Member,~IEEE,}\\ NaiQi Wu,~\IEEEmembership{Fellow,~IEEE,} and Mengchu Zhou,~\IEEEmembership{Fellow,~IEEE}

\thanks{}
\thanks{}
\thanks{M. Ghahramani is with Birmingham City University, UK, (e-mail: mohammadhossein.ghahramani@bcu.ac.uk)}

\thanks{Y. Qiao is with the Institute of Systems Engineering, Macau University of Science and Technology, Macau, (e-mail: yqiao@must.edu.mo)}
\thanks {N. Q. Wu is with Institute of Systems Engineering, Macau University of Science and Technology, Macau. (e-mail: nqwu@must.edu.mo).}
\thanks{M. C. Zhou is with the Helen and John C. Hartmann Department of Electrical and Computer Engineering, New Jersey Institute of Technology, Newark, NJ 07102, USA (e-mail: zhou@njit.edu).}

}

%
%

\markboth{}%
{Shell \MakeLowercase{\textit{et al.}}: Bare Demo of IEEEtran.cls for IEEE Journals}
%



\maketitle

\begin{abstract}
The integration of advanced technologies, such as Artificial Intelligence (AI), into manufacturing processes is attracting significant attention, paving the way for the development of intelligent systems that enhance efficiency and automation. This paper uses the term "Gentelligent system" to refer to systems that incorporate inherent component information (akin to genes in bioinformatics\textemdash where manufacturing operations are likened to chromosomes in this study) and automated mechanisms. By implementing reliable fault detection methods, manufacturers can achieve several benefits, including improved product quality, increased yield, and reduced production costs. To support these objectives, we propose a hybrid framework with a dominance-based multi-objective evolutionary algorithm. This mechanism enables simultaneous optimization of feature selection and classification performance by exploring Pareto-optimal solutions in a single run. This solution helps monitor various manufacturing operations, addressing a range of conflicting objectives that need to be minimized together. Manufacturers can leverage such predictive methods and better adapt to emerging trends. To strengthen the validation of our model, we incorporate two real-world datasets from different industrial domains. The results on both datasets demonstrate the generalizability and effectiveness of our approach.

\makeatletter{\renewcommand*{\@makefnmark}{}
\footnotetext{}\makeatother}
\end{abstract}


%
\IEEEpeerreviewmaketitle

\section{Introduction}\label{section.intro}
%
%
%
%

\IEEEPARstart{M}{ore} recently, manufacturing has embraced the Industrial Internet of Things (IIoT), where digital sensors, network technologies, and gentelligent components are integrated into manufacturing processes. A gentelligent component, as defined in the Collaborative Research Centre 653 project \cite{Denkena-1}, refers to components that intrinsically store information. The focus of that work is on encoding and preserving data within physical parts throughout the product lifecycle. 

Inspired by this concept, we extend the notion into what we define as a "\textit{gentelligent system}." In our model, we draw an analogy to biological systems, where manufacturing operations are viewed as chromosomes, and their associated features act like genes--each contributing specific behavioral traits to the overall system. Just as gene expression determines biological functions, the selected features influence the decision-making capabilities and reasoning and efficiency of our model. By integrating this perspective into a smart manufacturing context, we develop a system that adapts, learns, and evolves by selecting the most informative 'genes' (features) to optimize industrial performance. This interpretation not only reinforces the original idea of information-rich components, but also adds a novel layer of systemic intelligence and decision autonomy.

We are currently witnessing a significant transformation in manufacturing, known as Industry 4.0. This revolution involves maximizing production yields through IIoT innovations and integrating intelligent and autonomous systems powered by emerging technologies such as Artificial Intelligence (AI) and Machine Learning (ML). Additionally, Cloud and edge computing are enabling data storage and processing from virtually anywhere. AI has been successfully applied in various domains, including engineering \cite{2024-0,2023-1,2023-2}, urban planning \cite{ghahramaniSMC,ghahramaniIoT}, and manufacturing \cite {2023-3,2023-4,2023-5,2023-6,Ghahramani2020Smart1,SemiConductorGH,ChenWorkload2025,DengEvolutionary2025}, where it helps solve problems such as functional design, process planning, and product optimization. The application of AI in manufacturing can be categorized into decision support, data management, operations management, and life cycle management. The broad applicability of AI methods allows manufacturers to achieve numerous benefits, such as accurately estimating production costs, mitigating failures, and assessing the financial viability of proposed products.

Integrating data from various sensors with state-of-the-art AI technologies can lead to an efficient industrial ecosystem. However, this integration poses several challenges, necessitating practical solutions to address these issues. One of such critical challenges is the Value of Information (VoI) that we can extract from these data sources. The concept of VoI refers to the benefit derived from the industrial information obtained, particularly in decision-making processes. In the industrial context, not all data holds an equal value, and identifying the features that most effectively contribute to operational efficiency is critical. This is the focus of our study, which addresses challenges in smart manufacturing, particularly in optimizing feature selection and classification performance. Traditional optimization methods often struggle with the complexity and conflicting objectives inherent in these tasks. This is especially relevant for applications like predictive maintenance and fault detection, where enhanced model performance can reduce downtime, improve product quality, and decrease costs. Our solution provides a more scalable and efficient alternative to conventional methods, advancing AI-driven decision-making in manufacturing.

The challenge lies in accurately assessing and prioritizing the industrial features that can enhance predictive accuracy, reduce operational risks, or optimize resource allocation. Ensuring that AI models focus on high-value data can significantly improve outcomes, but this requires sophisticated methods to evaluate and integrate such data effectively. Addressing VoI is essential for maximizing the benefits of AI-driven decision-making in complex industrial systems.

\begin{figure*}
  \includegraphics[width=\textwidth]{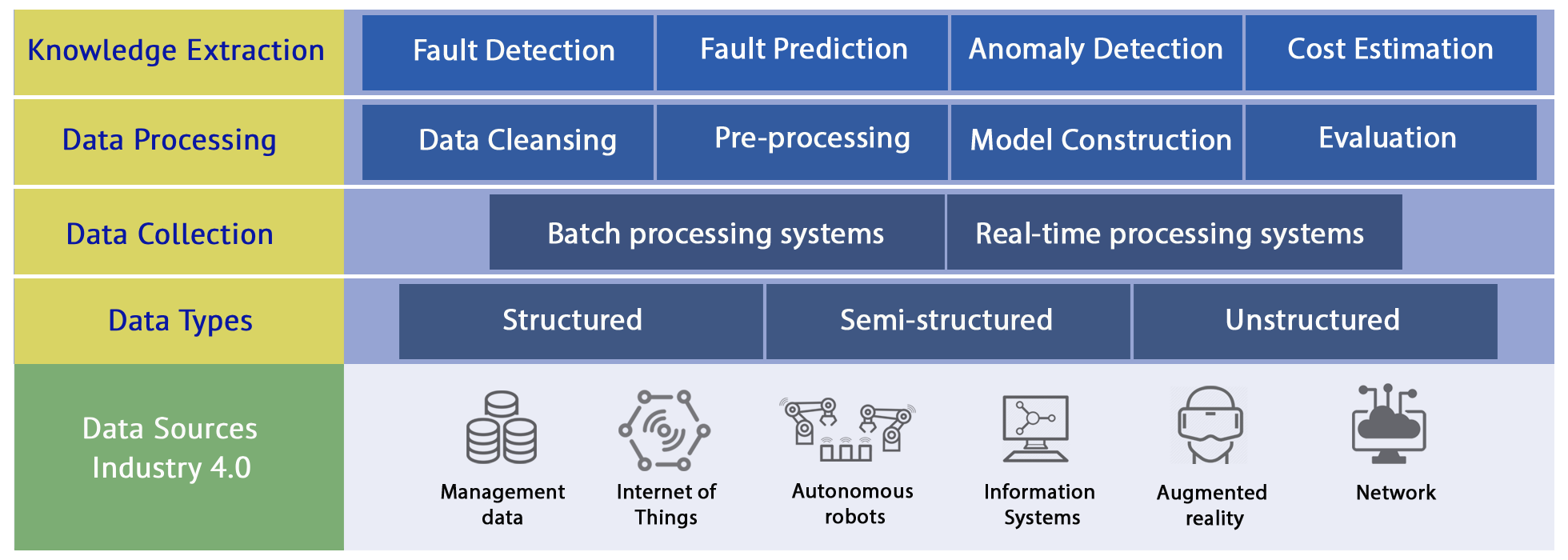}
  \caption{Different phases of manufacturing data life-cycle management.}
  \label{industry4}
\end{figure*}

Capturing industrial data is essential for detecting failures and minimizing system issues. However, to fully leverage this data, various processing operations are necessary throughout the product lifecycle to extract valuable insights. This systematic approach, involving multiple data processing steps (shown in Fig. \ref{industry4}), enables informed decision-making about priorities. The value of manufacturing data depends not only on its volume but also on the most informative features of the processes, as previously explained. Therefore, selecting or extracting these key features, either in batch mode or real-time, is crucial for maintenance, failure detection, prediction, and cost estimation. Analyzing feature dependencies can provide valuable insights for cost reduction, quality control, and inventory management. Motivated by these needs, this work proposes an AI-based model using an optimized evolutionary algorithm to study manufacturing processes and identify the most informative features.

Evolutionary algorithms enhance efficiency and reduce costs by finding global optima in the input space, while traditional methods like gradient-based techniques focus on local searches. In our previous work \cite{Ghahramani2020Smart1}, we proposed an AI-based model to mitigate cost and production risks, supporting the sustainable development of smart manufacturing. This model featured an intelligent, dynamic feature selection algorithm to reduce dimensionality and select the most informative features for model construction. We addressed challenges such as unbalanced classification and outliers, and developed a binary genetic algorithm to model production processes and determine the optimal number of features. A multi-layer perceptron was integrated to calculate the costs associated with different feature sets, which were then passed to various classification algorithms for comparison. A weighted-sum approach defined the cost function, breaking the feature selection problem into two scalar optimization subproblems, each solved using a binary genetic algorithm. It is worth mentioning that decomposition methods combine objective functions into a set of scalar problems that can be solved using a single optimization method, requiring multiple executions to achieve optimal solutions \cite{Fang2024-decop,Tan2024pso, Ming2024Const}. 

To reduce computational effort, we propose an optimized Multi-Objective Evolutionary Algorithm (MOEA) in this study. Unlike multiple runs in our previous work, this approach identifies a set of Pareto-optimal solutions in a single run, improving feature selection efficiency. The goal of this study is to implement an advanced model that examines production phases and extracts relevant manufacturing features by solving a Multi-Objective Optimization Problem (MOOP).

Recent decades have seen significant attention on MOEAs due to the rise of MOOPs. Real-world problems are often complex, involving multiple conflicting objectives that require careful evaluation of corresponding objective functions to optimize trade-offs. Various methods, particularly Pareto-based approaches that use dominance comparison mechanisms, have been developed to solve MOOPs. Non-dominated sorting methods, which incorporate both Pareto ranks and diversity criteria, have proven effective for robust optimization \cite{Hu.Efficient}, \cite{Wang.Evolutionary}. MOEAs are widely used in engineering, business, and medical fields. For example, Zhao et al. \cite{Zhao.Performance} used a multi-objective genetic algorithm to model a recirculating electric power steering system, validating results through testing. Djeffal et al. \cite{Djeffal.improved} proposed an optimization strategy to improve the electrical characteristics of analog circuits, with objectives based on voltage rate, transconductance, and conductance. Trivedi et al. \cite{Trivedi.Enhanced} tackled the unit commitment problem using an MOEA approach, focusing on minimizing costs and emissions.

Building on the effectiveness of the aforementioned approaches, we propose an AI-based method for manufacturing and process optimization. Identifying critical factors that influence manufacturing outcomes is essential for sustainability. Traditional fault detection methods often treat manufacturing data as singular values, limiting their ability to validate key operations or features. To overcome this, we introduce a novel hybrid feature selection method that integrates a two-level evolutionary algorithm with Deep Learning. This approach selects the most relevant manufacturing operations (features), which are then used as inputs for classifiers. The contributions of this work are as follows.

\begin{itemize}
\item We propose an optimal and dynamic strategy to study interconnected manufacturing flows by implementing a hybrid algorithm that combines a two-level MOEA sorting method with deep learning to model system non-linearity.

\item We improve our previous feature selection algorithm \cite{Ghahramani2020Smart1} to find all optimal solutions in a single run. This improvement significantly reduces computational expenses, making our approach more efficient.
\end{itemize}

The selected features are then passed to a classifier, and the error rate is measured. We evaluate the proposed model by comparing the results obtained in this work with some other MOEA methods. The remainder of this paper is organized as follows. Related work on feature selection methods is discussed in Section \ref{relatedWork}; Our research methodology is described in Section \ref{Method}; The proposed multi-objective approach is discussed in Section \ref{section.Model} along with the relevant discussion; The experimental settings and results are presented in Section \ref{Results}; Conclusions and future work are presented in Section \ref{section.conclusion}.

\section{Related Work}\label{relatedWork}
Dimensionality reduction refers to the process of selecting or extracting the most informative variables, ultimately reducing the number of input features. Two main approaches, feature extraction and feature selection, are used to achieve this. Feature extraction projects the original features into a new, lower-dimensional space, with Principal Component Analysis (PCA) and Linear Discriminant Analysis (LDA) being common examples. In contrast, feature selection involves choosing a subset of features that minimize redundancy while maximizing relevance to the target variable.

Much of the existing research has focused on linear PCA-based approaches for dimensionality reduction. These methods transform features into another space using linear combinations of the original features, which can make the results difficult to interpret in the original feature space \cite{Zhang-pca}. Additionally, PCA-based methods are limited by global linearity, making them less effective at capturing non-linear patterns. In such cases, non-linear methods, such as those based on kernel functions, may be more appropriate. However, interpreting patterns in these non-linear methods can be challenging because the new features lack a clear physical relationship with the original ones. To address issues of interpretability and readability, as well as the complexity introduced by feature extraction techniques, feature selection approaches can be employed. These methods are categorized into filter and wrapper techniques. Filter methods rank features and select them based on their relevance to the dependent variable, whereas wrapper methods evaluate features based on the predictive accuracy of a classification model.

In filter methods, both feature relevance and redundancy analyses are necessary. Identifying redundant features is particularly challenging because features with similar rankings may differ in relevance. Given the complexity of relevance analysis and the large number of features (nearly 600) in our case, a wrapper approach is more appropriate. An exhaustive search of all possible feature subsets is impractical, so global search methods, such as Evolutionary Computation (EC) techniques, are beneficial \cite{Ding2020Multi}. Derrac et al. \cite{Derrac-first} proposed a cooperative co-evolutionary algorithm for feature selection using Genetic Algorithms (GA), but their approach is stochastic and can yield different solutions based on initial conditions, leading to instability. Zamalloa et al. \cite{Zamalloa-Feature} developed a GA-based method for ranking features, but it may result in data loss and does not account for feature correlations. Multi-objective feature selection algorithms based on the Non-Dominated Sorting Genetic Algorithm II (NSGA-II) have been employed in \cite{Banerjee2007}. Furthermore, Gao et al. \cite{Gao2008feature} introduced a feature selection method using Ant Colony Optimization in network intrusion detection, leveraging the Fisher Discrimination Rate as heuristic information. Xue et al. \cite{Xue-PSO} proposed a multi-objective particle swarm optimization (PSO) approach, which includes two PSO-based algorithms--one using a non-dominated sorting approach and the other relying on distance measures, mutation, and dominance concepts. While effective, implementing two algorithms can be resource-intensive.

To tackle these challenges, we propose a two-level non-dominated sorting algorithm specifically designed for the feature selection task. This method incorporates a dominance phase, crowding distance mechanism, and DNN, making it well-suited for multi-objective optimization. As the number of objectives increases, especially beyond two, the search space becomes more complex and the proportion of non-dominated solutions grows, making it harder to distinguish the solutions from truly optimal ones. Our implementation of the Pareto dominance phase reduces the number of comparisons, helping to manage this complexity. The integrated DNN is used to estimate costs at various stages of the process. Note that alternative supervised neural networks, e.g., those in \cite{Gao-Wang,Gao-Zhou,Li-Zhou}, may also be employed.

\begin{figure}
  \includegraphics[width=\textwidth]{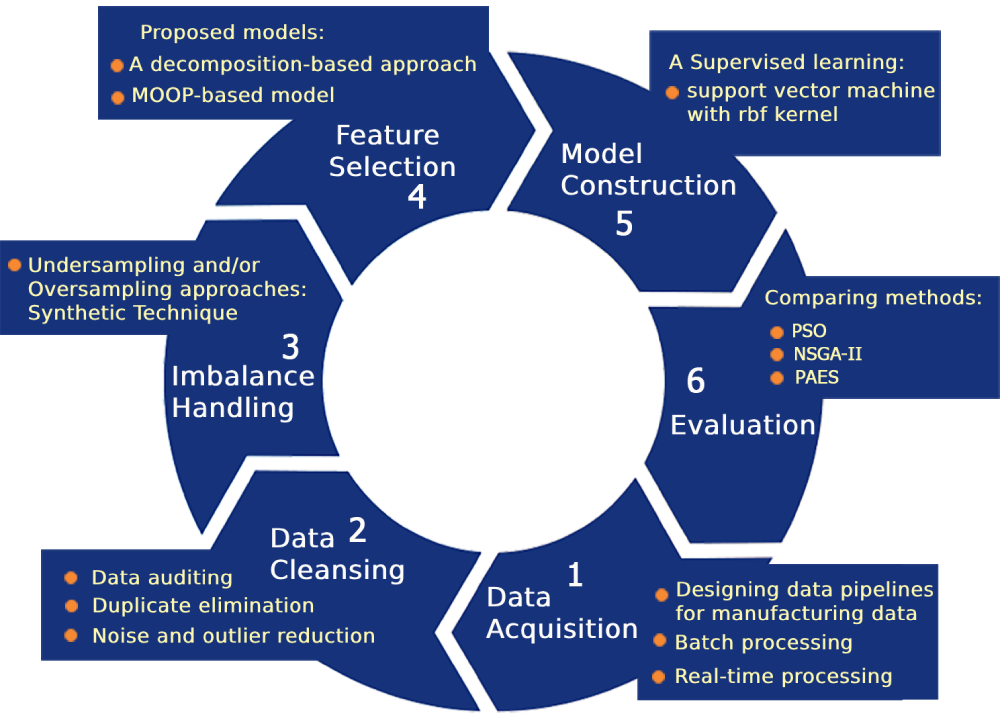}
  \caption{Different phases of the implemented model including the proposed feature selection methods.}
  \label{modelGraphic}
\end{figure}

\section{Proposed Approach}\label{Method}
The proposed approach involves multiple phases, including data acquisition, data cleansing, imbalance handling, feature selection, and model construction and evaluation (Fig. \ref{modelGraphic}). However, the primary focus is on enhancing the feature selection method presented in our previous work. For detailed discussions on data preprocessing (e.g., handling imbalances, noise reduction, and outlier elimination) and the classification model, we refer to \cite{Ghahramani2020Smart1} to conserve space. The goal of this paper is to explore the search space and identify all optimal solutions in a single run using a dominance-based approach. We address the technical aspects of implementing a Multi-Objective Evolutionary Algorithm (MOEA), such as solution sorting and evaluation, to achieve optimal solutions. The results of our proposed approach are compared with those of other models, including our previous model and different evolutionary algorithms.

\begin{figure}
  \includegraphics[width=\linewidth]{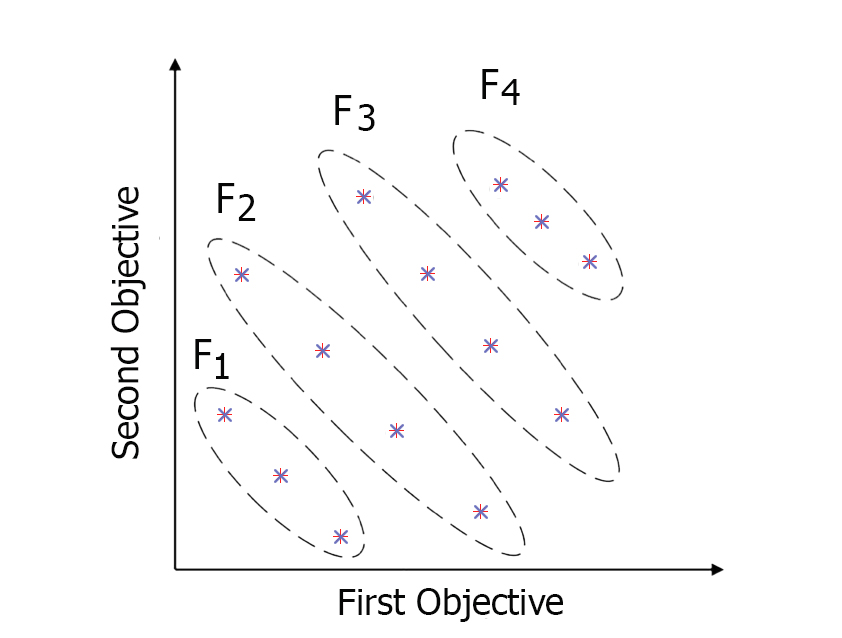}
  \caption{Representation of solutions assigned to different fronts.}
  \label{plotSorting}
\end{figure}

To maintain diversity in the solutions and facilitate comparison, the earlier approach required multiple runs. In contrast, our new model detects multiple Pareto-optimal solutions in a single run through a non-dominated sorting mechanism. Non-dominated sorting assesses solutions based on the dominance concept, assigning them to different fronts, as illustrated in Fig. \ref{plotSorting}. This ranking process divides the population into subsets, with ranks assigned by the sub-index $k$ of $F_k$.

We define the multi-objective problem as
 \begin{equation}
\min_{x} f(x)=(f_1(x),f_2(x), ..., f_m(x))
  \end{equation}
where $f(x)$ is a cost function comprising $m$ objective functions, and $x$ is a candidate solution, or a chromosome in Genetic Algorithm (GA) terminology, representing our feature space. Our objective is to determine the optimal number of features and reduce dimensionality, with the quality of solutions assessed based on trade-offs between conflicting objectives. For two solutions, $\alpha$ and $\beta$, from a multi-objective minimization problem, $\alpha$ is considered better than $\beta$ if:

 \begin{equation}
\forall i: f_i(\alpha) \leqslant f_i(\beta)  \text{   and   } \exists j: f_j(\alpha) < f_j(\beta)
  \end{equation}
where $i$ and $j$ $\in \{1,2,...,m\}$. 

In this work, the feature selection problem is formulated as a bi-objective optimization, aiming to minimize the number of features (first objective) and reduce the feature selection cost (second objective) simultaneously.

\subsection{Optimization Approach}
The datasets used in this study are publicly available and include the SECOM dataset and the Tennessee Eastman Process Simulation Data (TEP). These datasets can be obtained from multiple open repositories such as the UCI Machine Learning Repository, Kaggle, or other publicly accessible archives. Specific download links are not included here, as these datasets are hosted on public repositories whose URLs may change over time. SECOM, obtained from a semiconductor manufacturing facility, contains nearly 600 features (operation observations, i.e., wafer fabrication production data). The target feature is binomial (failure or success), indicating production status, and is encoded as 0 or 1. TEP, sourced from the Tennessee Eastman chemical production process, presents a distinct application scenario, with features related to fault detection and process optimization in a complex industrial setting. Like SECOM, its target feature is binary, indicating whether a fault was detected (failure or success). In our existing work, we have only used the SECOM dataset. The inclusion of TEP in this study helps us demonstrate the generalizability of our model across different industrial domains. Let $V=\{\nu_{1}, \nu_{2}, \ldots, \nu_{d}\}$ be the feature set and $L=\{$0$, $1$\}$ the target set, where, $0$ means failure, $1$ means success, and $d$ represents the dataset's dimensions. The matrix $X \in \mathbf{R}^{n \times d}$ (the original dataset) can be defined as follows:

 \begin{equation}
X=\begin{bmatrix}
         X_{1} \\   X_{2} \\ \vdots \\ X_{n}
   \end{bmatrix}
=\begin{bmatrix}
         x_{11} & x_{12} & \cdots  & x_{1d} \\   
         x_{21} & x_{22} & \cdots  & x_{2d} \\ 
         \vdots  & \vdots  & \ddots & \vdots  \\ 
         x_{n1} & x_{n2} & \cdots  & x_{nd} 
   \end{bmatrix}
  \end{equation}
where $\mathbf{R}$ is the real number set, $X_i$ (the $i^{\text{th}}$ input) is a \textit{d}-tuple, and $n$ is the number of observations. We define $Y$ as:

 \begin{equation}
Y=\begin{bmatrix}
         y_{1} ,   y_{2} , \ldots, y_{n}
        \end{bmatrix}^\top
  \end{equation}
where $y_{i}$ (i.e., the target value) is either $0$ or $1$, depending on whether it is relevant to the observation ($X_i$). Based on the given set of features, let $\Delta=\{\delta_{1}, \delta_{2}, \ldots, \delta_{n}\}$ be the predicted value for $n$ observations. We need to find a subset of features (based on both objective functions defined, $f: X^\prime \subseteq X \to \mathbf{R}_2$), such that the following criteria are met.

\begin{itemize}
  \item Set $|X^\prime|=k<d$. Find $X^\prime \subset X$, such that $f_1(X^\prime)$ is minimized: 
  
  \begin{equation} 
  f_1(X^\prime)= \frac{1}{n} \sum_{i=1}^{n} (Y_{i}-\Delta_{i})^2
  \label{MSE}
  \end{equation}
  
  \item Find optimal features while minimizing $|X^\prime|$.
\end{itemize}

\subsection{Key Characteristics}
Multi-objective evolutionary algorithms (MOEAs) can be categorized into decomposition, indicator, and dominance-based methods. In the past study, we proposed a decomposition method (a weighted sum technique) based on a binary GA that involved repeated sequences of operations such as parent selection, recombination, and mutation. In each iteration, cost-efficient chromosomes (features) from the population were sorted and selected. This method decomposed the feature selection problem into several scalar optimization sub-problems, using a weighted sum approach to select optimal features.

However, to simultaneously optimize both objectives, the decomposition method is not suitable, as it does not allow for sorting solutions in the same manner. Therefore, this paper addresses the optimization problem using a Pareto-dominance selection principle. The goal is to identify a set of optimal solutions using a dominance-based MOEA technique. Each solution is compared with others, and those not dominated by others are assigned to front $F1$. The remaining solutions are then compared, with non-dominated solutions assigned to front $F2$. This process continues until all solutions are assigned. In the second phase of the algorithm, solutions within each front are sorted based on a crowding distance measure, ensuring good diversity among the population's solutions. This method allows for the detection of all non-dominated solutions, with the best one selected.

\subsection{Feature Selection Model}\label{section.Model}
As noted, the problem is framed as a bi-objective problem. The defined objectives—minimizing the number of features and the feature selection cost-function—must be optimized simultaneously, requiring a trade-off between them. Earlier, we addressed the problem of finding the optimal number of features using a decomposition-based strategy involving a set of weights. Each observation, comprising different subsets of manufacturing operations/features with varying dimensionalities, was treated as an individual. Individuals were sorted based on their corresponding cost and then selected. However, with two objectives to optimize, that approach is inadequate, as objectives cannot be compared directly. Our current solution, which deals with only two objective functions, avoids multiple objective comparisons, making it an efficient and computationally inexpensive approach.

\begin{figure}
  \includegraphics[width=\linewidth]{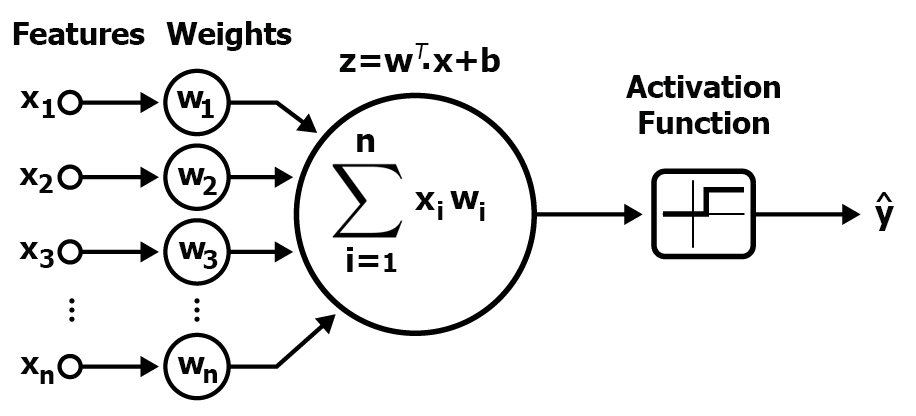}
  \caption{The neural network architecture considered to compute feature selection costs.}
  \label{neuron}
\end{figure}

In this work, we propose a multi-objective model based on a Pareto-dominance mechanism, combining a non-dominated sorting GA with a DNN. The integrated DNN has a feed-forward architecture consisting of an input layer, one hidden layer, and an output layer. The input layer size corresponds to the number of selected features in a given chromosome. The hidden layer includes 15 neurons, a value derived from empirical analysis and tuning. The output layer contains a single node representing the predicted class. We use the leaky ReLU activation function in the hidden layer and a sigmoid activation function in the output layer, suitable for binary classification. The network is optimized by using a momentum-based optimizer \cite{Wen-Zhou} with a learning rate of 0.001. We trained DNN for 10 epochs in each iteration to limit computational cost. The results are presented in the model comparison section. As shown in Fig. \ref{neuron}, the network receives different sets of features as input and calculates their corresponding costs. A weighted average of input values (using weight vectors and a bias unit, $z=w_1x_1+w_2x_2+...+w_nx_n+b=w^Tx+b$), is computed and passed through a nonlinear activation function. Errors are then calculated using the Mean Squared Error function (Eq. \ref{MSE}) and the Cross-Entropy Loss function (Eq. \ref{Loss}).

\begin{equation}
Loss(\hat{y_i},y_i)= -(y_i\textbf{\textit{log}}(\hat{y_i})+(1-y_i)\textbf{\textit{log}}(1-\hat{y_i}))
\label{Loss}
\end{equation}
where $\hat{y_i}$ is the predicted value for the $i^{\text{th}}$ observation. In each iteration, the network was trained for 10 epochs, with the average cost considered as the output for that iteration. This phase aimed to calculate the costs associated with different feature sets, and the use of 10 epochs was intended to mitigate under-fitting concerns. Increasing the number of epochs could extend computing time, given the iterative nature of the optimization algorithm.

A non-dominated sorting GA, similar to the binary GA used in the previous work, includes phases like Parent Selection, Crossover, and Mutation to create a new population. For more details on population creation and discussions on exploration and exploitation in evolutionary algorithms, refer to \cite{Ghahramani2020Smart1}. It is important to note that the method for sorting and selecting solutions in this work differs from the previous study. In this work, exploration is encouraged through the use of a crowding distance measure during the non-dominated sorting process, which ensures that a diverse set of solutions is maintained across generations. By preserving individuals that are well-distributed across the objective space, the algorithm is able to explore a wide range of possible solutions and avoid premature convergence to suboptimal regions of the search space.
However, exploitation is handled by the elitism strategy in the selection process, where the best-performing individuals are always carried forward to the next generation. This ensures that high-quality solutions are preserved while new generations explore new areas of the solution space. The adaptive mutation rate also contributes to the balance, allowing for finer adjustments in the solution space as the algorithm progresses, enabling more focused exploitation in later generations. By combining these techniques, i.e., crowding distance for exploration, elitism for exploitation, and adaptive mutation for fine-tuning, we achieve a balance that prevents stagnation and ensures the algorithm efficiently converges to high-quality solutions without losing diversity.

Addressing a multi-objective problem involves finding a set of Pareto-optimal solutions. In this work, solutions are ranked using a two-level scheme: a non-dominated sorting mechanism and crowding distance. The non-dominated sorting arranges individuals according to the Pareto dominance principle, while crowding distance ranks solutions based on their contribution to diversity. The following section explains this selection scheme, and the pseudo-code for the feature selection model based on the dominance-based approach is provided in Algorithm \ref{featureSelection1}.

\begin{algorithm}[h]
\SetAlgoLined
\SetKwInOut{Input}{Input}
\SetKwInOut{Output}{Output}

\Input{
    $X$ (the dataset)\;
    $n = |X|$ (the number of features)\;
    Cost Function (implemented using DNN)\;
}

\Output{
    Optimal features\;
}

$\eta_p \gets \text{number of population}$\ {\scriptsize\tcp*{Population size}}

$\eta_r \gets \text{offspring rate}$\ {\scriptsize\tcp*{Rate at which offspring are generated}}

Initialize population $\rho_0$\ {\scriptsize\tcp*{Initial solutions}}

Sort population:\\
\quad First level ranking (Non-dominated Sorting, Alg. 2)\ {\scriptsize\tcp*{Initial sorting based on dominance}}

\quad Second level ranking (Crowding Distance, Alg. 3)\ {\scriptsize\tcp*{Sorting based on crowding distance to maintain diversity}}

\While{termination condition is not met}{
    $Q_t \gets \{\}$ {\scriptsize\tcp*{Offspring and mutants list initialization}}
    \For{$i \gets 1$ \KwTo $\eta_r$}{
        Select two parents $P_1$ and $P_2$\ {\scriptsize\tcp*{Selection of parents for crossover}}
        $\omega(i) \gets \text{CrossoverFn}(P_1, P_2)$\ {\scriptsize\tcp*{Crossover to create offspring}}
        $\mu(i) \gets \text{MutationFn}(\omega(i))$\ {\scriptsize\tcp*{Mutation of offspring}}
        $Q_t \gets Q_t \cup \{\omega, \mu\}$\ {\scriptsize\tcp*{Add offspring and mutants to list}}
    }
    Select $\eta_p$-best individuals based on:\
    \quad Non-dominated Sorting and Crowding Distance\ {\scriptsize\tcp*{Select top individuals based on Pareto-optimality and diversity}}
}

Output the result\;
\caption{Pseudo-code for the feature selection model based on MOEA.\\
$\eta_p$ - Population size \\
$\eta_r$ - Offspring rate \\
$\rho_0$ - Initial population of solutions \\
$Q_t$ - List of offspring and mutants}
\label{featureSelection1}
\end{algorithm}

\subsection{Ranking schema: first phase} 
Each solution to the feature selection problem is defined by two objective functions, $f_1$ and $f_2$, as specified earlier. The goal is to identify a group of ideal solutions, known as non-dominated solutions. Assume $P$ is a set of solutions where no two members dominate each other; We can say $s_1$ dominates $s_2$ only if $s_1$ is not worse than $s_2$ in all objectives:

\begin{equation*}
f_{i}(s_1) \leq f_{i}(s_2)\text{,       }\forall \text{ } i \in \{1, 2\}
\end{equation*}

Additionally, at least one of $s_1$'s objective values must be superior to $s_2$, i.e.

\begin{equation*}
\begin{multlined}
\exists \text{ } i\in \{1,2\}, \text{ } \ni f_{i}(s_1) <  f_{i}(s_2)\\ 
\end{multlined}
\end{equation*}

The ranking process uses Pareto orders. During the non-dominated sorting phase, all solutions are divided into subsets (fronts) based on non-dominance comparisons. Let $F= F_1, F_2,..., F_k$ represent a set of $k$ Pareto fronts where $F_i$ represent the $i^\text{th}$ Pareto front containing $m$ solutions. The goal at this stage is to find Pareto-optimal solutions. For each solution $s_p$, two entities are computed: $n_p$, the number of solutions that dominate $s_p$, and $\check{S}_p$, the set of solutions dominated by $s_p$. Solutions that are not dominated (i.e., where $n_p=0$) are categorized as $F_1$ when all solutions are compared. For each solution $s_q$ in $\check{S_p}$, the corresponding 
$n_q$ value is decreased by one. Members of $F_2$ are solutions whose values are $0$. The process continues until all solutions are assigned to fronts. The pseudo-code for this process is provided in Algorithm \ref{non_dominated_sorting}. Note that solutions with the same rank are further ordered by crowding distance, which is discussed in the next section.

\begin{algorithm}[h]
\SetAlgoLined
\SetKwInOut{Input}{Input}
\SetKwInOut{Output}{Output}

\Input{
    Population set\;
    Cost Function\;
}

\Output{
    Sorted population and Pareto Fronts\;
}

$F \gets \{\}$  {\scriptsize \tcp*{Initialize Pareto fronts}}

$\check{S_p} \gets \{\}$ {\scriptsize\tcp*{Initialize dominated solutions}}

$n_p \gets 0$  {\scriptsize\tcp*{Dominance count for each solution}}

\For{\text{each $s_p$ \label{for_sp}}}{
    \For{\text{each $s_q$ \label{for_sq}}}{
        \uIf{$s_p \prec s_q$ {\scriptsize\tcp*{$s_p$ dominates $s_q$ }}}{
            $\check{S_p} \gets \check{S_p} \cup \{s_q\}$ \label{add_to_Sp}\;
        }
        \uElseIf{$s_q \prec s_p$ {\scriptsize\tcp*{$s_q$ dominates $s_p$ }}}{
            $n_p \gets n_p + 1$\;
        }
    }
    \If{$n_p = 0$ \label{pareto_front_check}}{
        $F_1 \gets F_1 \cup s_p$ \label{add_to_F1}\;
    }
}

$i \gets 2$ \label{init_i}\;
Initialize index for the next Pareto front

\While{$F_i \neq 0$ \label{while_next_front}}{
    $ \Omega \gets \{\}$ {\scriptsize\tcp*{List to maintain solutions}} 
    (Temporary list for the next Pareto front)

    \For{\text{each $s_p \in F_i$ \label{iterate_Fi}}}{
        \For{\text{each $s_q \in \check{S_p}$ \label{iterate_Sp}}}{
            $n_q \gets n_q - 1$ \label{decrement_n_q}\;
            \If{$n_q = 0$ \label{pareto_select}}{
                $\Omega \gets \Omega \cup \{s_q\}$ \label{add_to_Omega}\;
            }
        }
    }
    $i \gets i + 1$ \label{increment_i}\;
    $F_i \gets \Omega$ \label{assign_to_Fi}\;
}

\Output{Sorted population and Pareto fronts \label{final_pareto_fronts};}
\caption{Pseudo-code for first-level ranking using non-dominated sorting. \\
$F$ - Pareto fronts set, storing non-dominated solutions \\
$\check{S_p}$ - Set of dominated solutions \\
$n_p$ - Dominance count for each solution \\
$s_p, s_q$ - Solutions to be compared \\
$i$ - Index for the current Pareto front \\
$\Omega$ - Temporary list for storing solutions for the next Pareto front}
\label{non_dominated_sorting}
\end{algorithm}

\subsection{Ranking schema: second phase} 
\begin{figure}
  \includegraphics[width=\linewidth]{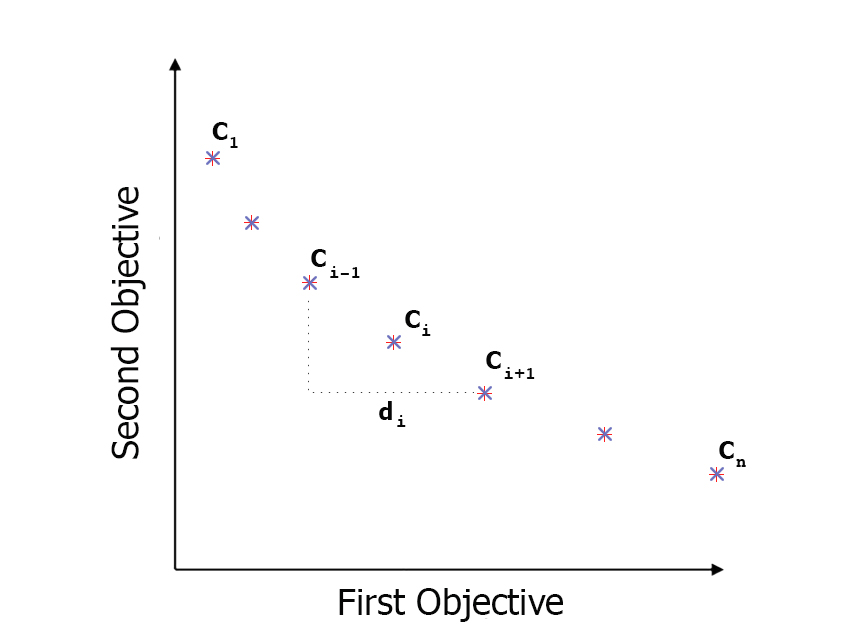}
  \caption{Representation of solutions in the second phase}
  \label{crowdingDistance}
\end{figure}

Crowding distance is used to estimate similarities among solutions at different fronts, arranging them in ascending order using this density-estimation metric. The average distance from each reference point to its two adjacent solutions is measured, which calculates the relative isolation of each solution using the comparison operator mentioned earlier. Let 
$C_i$ be the cost associated with the $i^\text{th}$ solution, as shown in Fig. \ref{crowdingDistance}. The average distance is calculated based on crowding distance, with similarities between solutions estimated and arranged accordingly.

Based on crowding distance, similarities among solutions at different fronts are estimated. The solutions are then arranged in ascending order according to this density-estimation metric. For each reference point, the average distance from its two adjacent solutions is measured. Consequently, the relative isolation of each solution is determined using the aforementioned comparison operator. Let $C_i$ denote the cost associated with the $i^\text{th}$ solution, as illustrated in Fig.~\ref{crowdingDistance}. The average distance can thus be computed based on the crowding distance, as follows:

\begin{equation*}
d(i) =\frac{C_{i+1}-C_{i-1}}{|C_1-C_n|} 
\end{equation*}

This evaluation is performed across all solutions within a particular front selected in the previous stage, with distance measures computed for each solution in the objective space. This process enables the sorting of all solutions within the given front. It is essential to prioritize boundary solutions such as $C_1$ and $C_n$, as they play a critical role in maintaining diversity and should be preserved throughout the evaluation. The pseudo-code for this second phase of ranking is presented in Algorithm~\ref{CrowdingDist}. Final solutions are selected after iteratively executing all steps in both ranking processes and evaluating the trade-offs among conflicting objectives. This decision should be made after considering other Pareto-optimal solutions.

\begin{algorithm}[h]
\SetAlgoLined
\SetKwInOut{Input}{Input}
\SetKwInOut{Output}{Output}

\Input{
    Population set ($\rho$)\;
    Pareto Fronts\;
    Cost Function\;
}

\Output{
    Sorted population\;
}

$n \gets |\rho|$\ {\scriptsize\tcp*{Total number of solutions}}

$n_F \gets \text{Number of Pareto fronts}$\ {\scriptsize\tcp*{Total number of Pareto fronts}}

$n_o \gets \text{Number of objectives}$\ {\scriptsize\tcp*{Number of objectives}}

$d \gets \text{Dataframe with $n$ rows and $n_o$ columns}$\ {\scriptsize\tcp*{Dataframe to store crowding distances}}

\For{$k \gets 1$ \KwTo $n_F$}{
    \text{Calculate costs of each member of $F(k)$}\
    $n_k \gets \text{Number of members of $F(k)$}$\ {\scriptsize\tcp*{Number of solutions in front $F(k)$}}

    \For{$j \gets 1$ \KwTo $n_o$}{
        $C \gets \text{costs for the $j^\text{th}$ objective}$\;
        \text{Sort } $C$\;
        $d(1) \text{ and } d(n) \gets \infty$\ {\scriptsize\tcp*{Assign infinity to boundary solutions}}

        \For{$i \gets 2$ \KwTo $n-1$}{
            $d(i) \gets \frac{C_{i+1} - C_{i-1}}{|C_1 - C_n|}$\ {\scriptsize\tcp*{Calculate crowding distance for each solution}}
        }

        \For{$i \gets 1$ \KwTo $n_k$}{
            $\text{Crowding distance for the $i^\text{th}$ member of $F(k)$} = \sum_{i=2}^{n-1} d(i)$\ {\scriptsize\tcp*{Sum up crowding distances to ensure diversity}}
        }
    }
}

\Output{Sorted population\;}
\caption{Pseudo-code for the second-level ranking schema using tailored crowding distance.\\
$F$ - Pareto fronts set, storing non-dominated solutions \\
$d$ - Dataframe storing crowding distances for each solution \\
$n_k$ - Number of solutions in the $k^{\text{th}}$ Pareto front \\
$C$ - Costs associated with the $j^{\text{th}}$ objective \\
$d(i)$ - Crowding distance for the $i^{\text{th}}$ solution \\
$n$ - Total number of solutions in the population}
\label{CrowdingDist}
\end{algorithm}

\section{Results}\label{Results}
As discussed, the proposed algorithm for our feature selection problem is based on an evolutionary multi-objective optimization combined with a DNN. It aims to identify a diverse set of Pareto-optimal solutions by evaluating different subsets of features, selecting those that best align with our optimization criteria. The DNN is integrated to calculate costs using our predefined cost function, with the number of neurons and layers determined through trial and error. Additionally, the volume and dimensionality of our dataset were taken into account when defining the initial population rate. For more information on setting various parameters in evolutionary computing methods, please refer to \cite{Ghahramani2020Smart1, Gao2019Dendritic}.

\begin{figure}
  \includegraphics[width=\linewidth]{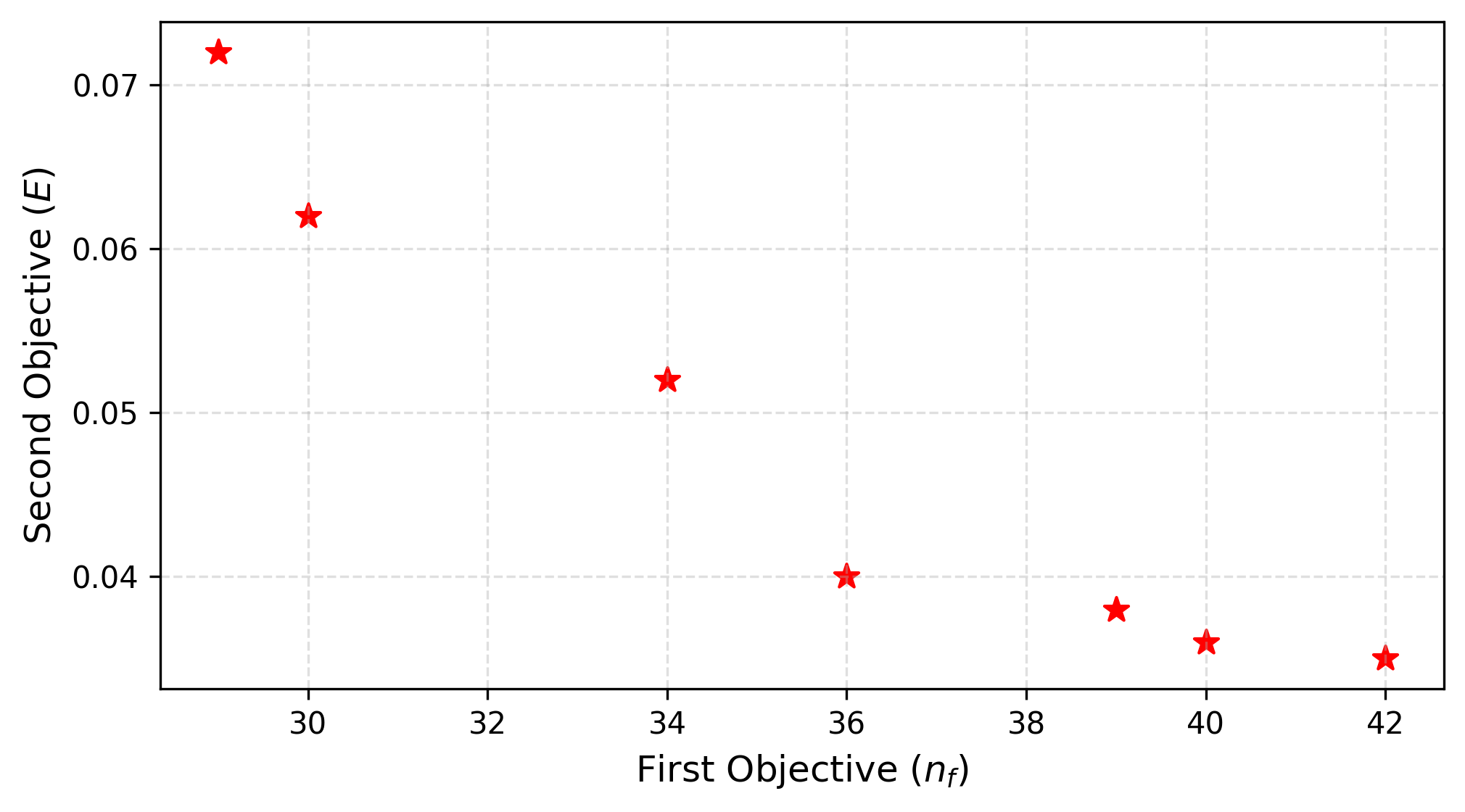}
    \caption{Non-dominated members identified based on the two objectives: number of features $(n_f)$ and error rate $(E)$. This figure shows the results for the SECOM dataset, where the optimal solution with 36 features is selected. For the TEP dataset, 38 features are selected as the optimal solution.}
  \label{featureSelection}
\end{figure}

Our multi-objective optimization method generates various subsets of features, leading to a diverse set of Pareto-optimal solutions after a series of iterative computations. We defined our feature selection task as an optimization problem with two conflicting objectives. To address this, we established a dominance relation among solutions, ranking them accordingly. This ranking process resulted in different fronts. Subsequently, we considered crowding distance for secondary ranking to select the best solution within a specific front. Figure \ref{featureSelection} shows the achieved optimal solutions (for the SECOM dataset) that are not dominated by others in the process, constituting our Pareto-optimal set. Among these, the solution with $36$ was selected as our optimal choice, as it resulted in the highest classification accuracy. For the TEP dataset, $38$ features were selected as the optimal solution based on the same criteria.

\begin{figure}
  \includegraphics[width=\linewidth]{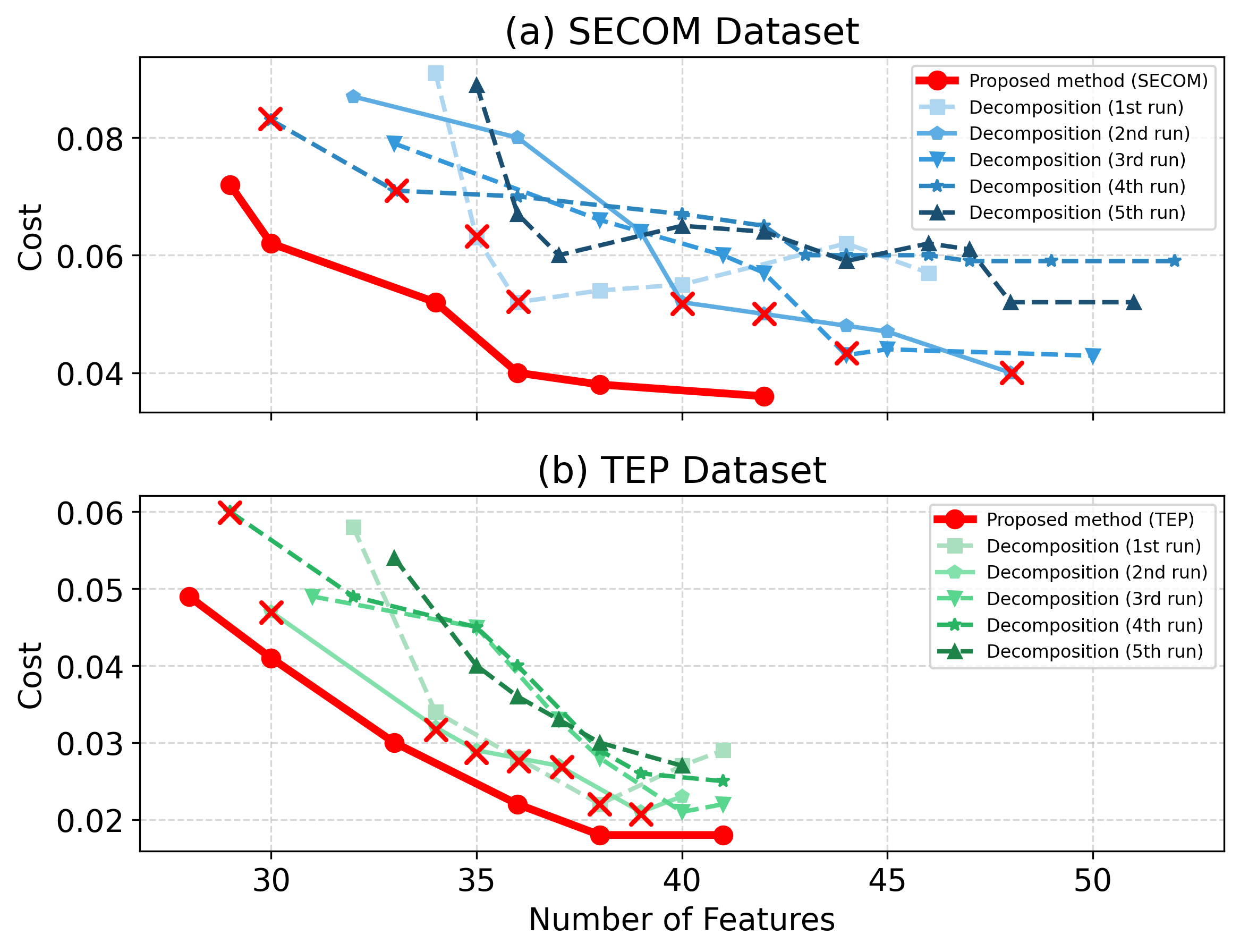}
    \caption{{Comparison of the proposed multi-objective feature selection method with decomposition-based approaches on two datasets: (a) SECOM and (b) TEP. Pareto-optimal points from five runs of the decomposition method are marked with red crosses. The proposed model consistently achieves lower cost with fewer features, confirming its generalizability. The red line demonstrates the proposed approach versus running the decomposition method multiple times.}}
  \label{comp3}
\end{figure}

Figure \ref{comp3} illustrates the results of implementing the decomposition method compared to our proposed method. As noted earlier, the decomposition method requires several runs to achieve Pareto-optimal solutions, whereas our method can provide optimal results in just a single run. In Figure \ref{comp3}, the decomposition method was performed five times, with the resulting Pareto-optimal solutions marked with red multiplication signs.

\begin{figure}
  \includegraphics[width=\linewidth]{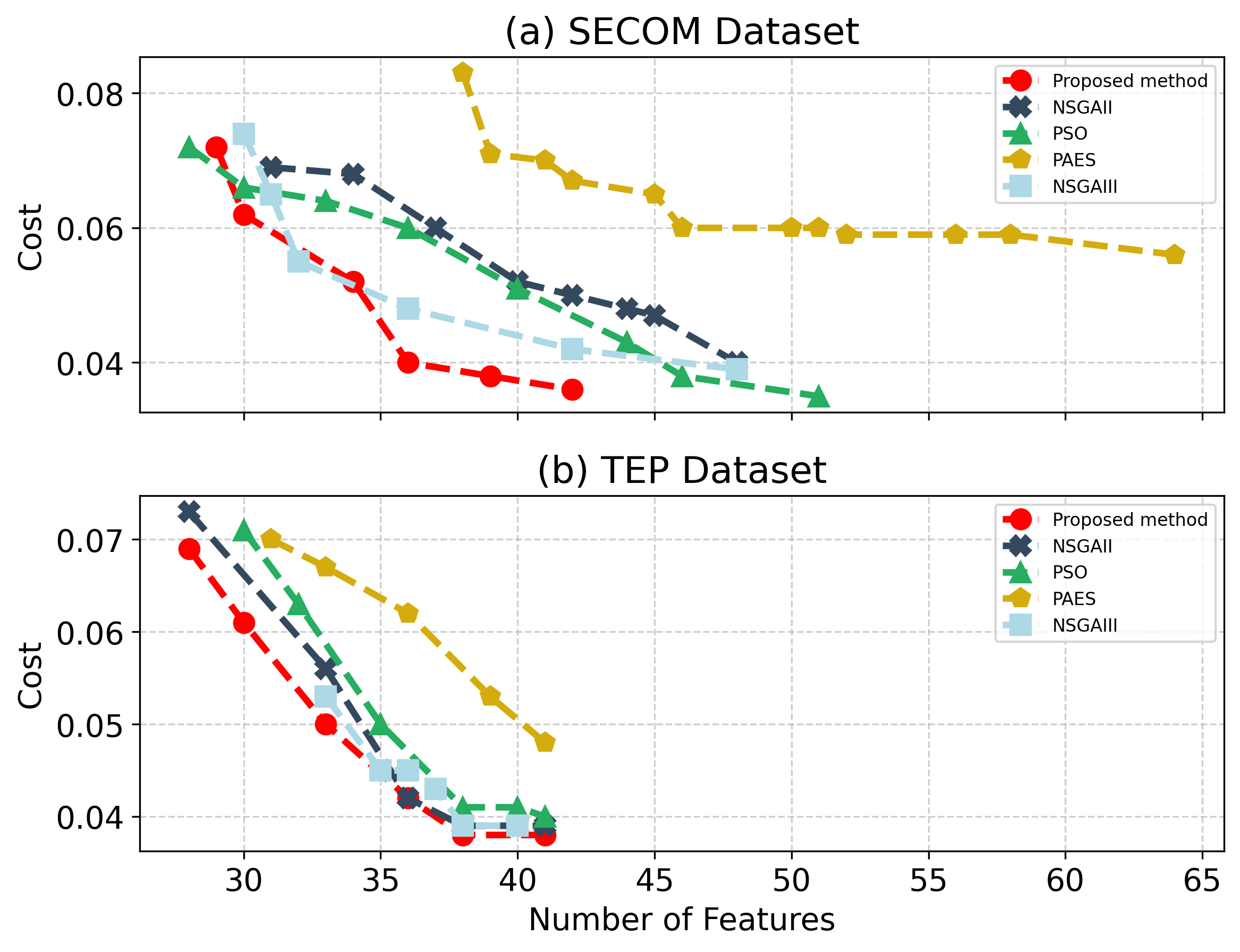}
\caption{{Comparison of the proposed feature selection approach with other evolutionary algorithms: (a) For the SECOM dataset (~600 features), the proposed method consistently achieves the lowest cost while selecting fewer features, demonstrating its effectiveness in high-dimensional settings. (b) For the TEP dataset (41 features), some models perform comparably; however, the proposed approach still exhibits superior overall performance.}}
  \label{comp1}
\end{figure}

To evaluate the performance of our proposed approach, we compared it against other evolutionary algorithms, including Particle Swarm Optimization (PSO), NSGA-II, NSGA-III, and Pareto Archived Evolutionary Strategy (PAES). The comparisons are shown in Figure \ref{comp1}, where the x-axis represents the number of features, and the y-axis represents the associated costs of the tested methods. As shown, the proposed model algorithm overall outperforms the others. It should be noted that PSO,  NSGA-II, and NSGA-III are based on the dominance concept, whereas PAES utilizes a local search strategy \cite{Jamwal2016Multicriteria}. These methods face challenges in balancing exploration and exploitation. NSGA-II, for example, relies on non-dominated sorting to maintain diversity, but as the dimensionality increases, it becomes computationally expensive and may fail to explore the full range of possible solutions. Also, PSO does not maintain population diversity, resulting in premature convergence to suboptimal solutions, especially in high-dimensional space \cite{Gad-Houssein}.

For the classification experiment, we split the dataset into 70\% for training and 30\% for testing. The training process was conducted, and the selected features from different feature selection algorithms were evaluated using the test set. We then calculated the classification error rate for each of the tested evolutionary algorithms. In this phase, we used a Support Vector Machine (SVM) algorithm to assess classification accuracy, as it performed well in \cite{Ghahramani2020Smart1}. The SVM was applied to all feature subsets selected by the different algorithms.

As shown in Table \ref{com-clf}, the proposed model outperforms all other methods on both datasets. On SECOM dataset, it achieves a classification accuracy of 94.6\% with 36 features, and on TEP, it reaches 95.8\% accuracy with 38 features. PSO follows closely, with 93.4\% accuracy on SECOM and 95.2\% on TEP, using 44 and 38 features, respectively. NSGA-II performs slightly worse than PSO, with accuracies of 92.2\% on SECOM and 95\% on TEP, using 42 features on SECOM and 36 on TEP. NSGA-III's performance is similar to NSGA-II's, achieving 93.7\% on SECOM and 93\% on TEP with 42 and 37 features, respectively. PAES consistently shows the lowest accuracy, with 91\% on SECOM and a significant drop to 85\% on TEP. It is important to note that while both the proposed model and PSO consider 38 features on the TEP dataset, they differ in 3 features. The proposed model identifies the most informative features, resulting in higher classification accuracy compared to PSO. These results demonstrate the effectiveness of the proposed model, achieving high classification accuracy with fewer features than its peers, especially in high-dimensional data settings.

\begin{table}[h!]
\centering
\caption{Comparison of Different Models on SECOM and TEP Datasets}
\begin{tabular}{lcc|cc}
\toprule
\multirow{2}{*}{\textbf{Models}} & \multicolumn{2}{c|}{\textbf{Num. Features}} & \multicolumn{2}{c}{\textbf{Accuracy}} \\
\cmidrule(lr){2-3} \cmidrule(lr){4-5}
 & SECOM & TEP & SECOM & TEP \\
\midrule
Proposed  & 36 & 38 & 94.6\% & 95.8\% \\
PSO       & 44 & 38 & 93.4\% & 95.2\% \\
NSGA-II   & 42 & 36 &  92.2\%  & 95\%   \\
NSGA-III  & 42 & 37 &   93.7\%    & 93\%   \\
PAES      & 52 & 36 & 91\% & 85\%   \\
\bottomrule
\end{tabular}
\label{com-clf}
\end{table}

Taguchi’s experimental design method and variance analysis were also employed to determine the optimal combination of user-defined parameters for the tested algorithms \cite{Gao2019Dendritic}. The maximum number of iterations for all the multi-objective algorithms was set to 100. For PSO, the maximum velocity was chosen as 0.7, and the population size was set at 400. We also configured the inertia weight, personal learning coefficient, and global learning coefficient based on constriction coefficients. In the decomposition method, the crossover rate was set to 0.8, the mutation rate to 0.3, and the population size to 700. As mentioned earlier, the decomposition method requires multiple runs to obtain a single solution from each run. In contrast, multi-objective methods like the proposed method, PSO, NSGA-II, NSGA-III, and PAES yield a set of non-dominated solutions. Table \ref{Params} outlines the various parameter settings used in the experiments for the proposed model.

\begin{table}[ht]
\caption{{Comparing our hybrid model across parameter settings. The lowest cost for SECOM is 0.0363, and for TEP it is 0.0283. An Offspring Rate of 0.6 yields higher costs, while rates of 0.7 and 0.9 produce intermediate results.}}

\centering 
\scriptsize
\begin{tabular}{l c c r r} 
\hline\hline 
 Offspring Rate & Population & Num. Neurons & Cost (SECOM) & Cost (TEP) \\ [0.5ex]
\hline 
 &  
 &10 & $0.0971$ & 0.0812   \\[-0.5ex]
{0.6} & {600}
& 15 & $0.092$  & 0.805   \\[-0.5ex]
 &  
 &20 & $0.0863$  & 0.0819    \\[0.5ex]
\hline 
 & 
 &10 & $0.0903$  & 0.0728   \\[-0.5ex]
{0.6} & {700}
& 15 & $0.0893$  & 0.0731   \\[-0.5ex]
 &
 &20 & $0.0899$  & 0.0719    \\[0.5ex]
\hline 
 & 
 &10 & $0.0822$  & 0.0754   \\[-0.5ex]
{0.6} & {800}
& 15 & $0.0803$  & 0.0743   \\[-0.5ex]
 & 
 &20 & $0.0916$  & 0.0792    \\[0.5ex]
\hline 
 &
 &10 & $0.0837$  & 0.0840   \\[-0.5ex]
{0.6} & {900}
& 15 & $0.0812$  & 0.0812   \\[-0.5ex]
 & 
 &20 & $0.1053$  & 0.0827    \\[0.5ex]
\hline 
 &
 &10 & $0.0598$  & 0.0625   \\[-0.5ex]
{0.7} & {600}
& 15 & $0.0551$  & 0.0655   \\[-0.5ex]
 & 
 &20 & $0.056$   & 0.0663    \\[0.5ex]
\hline 
 & 
 &10 & $0.0562$  & 0.0620   \\[-0.5ex]
{0.7} & {700}
& 15 & $0.0538$  & 0.0631  \\[-0.5ex]
 & 
 &20 & $0.0527$  & 0.0624    \\[0.5ex]
\hline 
 & 
 &10 & \textbf{$0.0392$} & 0.0463   \\[-0.5ex]
{0.7} & {800}
& {15} & \textbf{0.0363} & 0.0445   \\[-0.5ex]
 & 
 &20 & $0.0386$  & 0.0471    \\[0.5ex]
\hline 
 & 
 &10 & $0.0471$  & 0.0442   \\[-0.5ex]
{0.7} & {900}
& 15 & $0.0469$  & 0.0450   \\[-0.5ex]
 & 
 &20 & $0.0466$  & 0.0451    \\[0.5ex]
\hline 
 &
 &10 & $0.0771$  & 0.0376   \\[-0.5ex]
{0.8} & {600}
& 15 & $0.0762$  & 0.0360   \\[-0.5ex]
 & 
 &20 & $0.0766$  & 0.0384    \\[0.5ex]
\hline 
 & 
 &10 & $0.0753$  & 0.0301    \\[-0.5ex]
{0.8} & {700}
& 15 & $0.0761$  & \textbf{0.0283}    \\[-0.5ex]
 & 
 &20 & $0.0744$  & 0.0293    \\[0.5ex]
\hline 
 & 
 &10 & $0.0698$  & 0.0370   \\[-0.5ex]
{0.8} & {800}
& 15 & $0.0652$  & 0.0368     \\[-0.5ex]
 & 
 &20 & $0.0681$  & 0.0369    \\[0.5ex]
\hline 
 & 
 &10 & $0.0711$  & 0.0372    \\[-0.5ex]
{0.8} & {900}
& 15 & $0.0702$  & 0.0373    \\[-0.5ex]
 & 
 &20 & $0.0728$  & 0.0374     \\[0.5ex]
\hline 
 & 
 &10 & $0.0803$  & 0.0578    \\[-0.5ex]
{0.9} & {600}
& 15 & $0.079$   & 0.0580    \\[-0.5ex]
 &
 &20 & $0.0826$  & 0.0581    \\[0.5ex]
\hline 
 & 
 &10 & $0.0744$  & 0.0575   \\[-0.5ex]
{0.9} & {700}
& 15 & $0.0723$  & 0.0576  \\[-0.5ex]
 &
 &20 & $0.076$   & 0.0578    \\[0.5ex]
\hline 
 &
 &10 & $0.0811$   & 0.0579    \\[-0.5ex]
{0.9} & {800}
& 15 & $0.0827$   & 0.0580    \\[-0.5ex]
 &
 &20 & $0.0825$   & 0.0581     \\[0.5ex]
\hline 
 & 
 &10 & $0.0800$   & 0.0578    \\[-0.5ex]
{0.9} & {900}
& 15 & $0.0781$   & 0.0577    \\[-0.5ex]
 &
 &20 & $0.0779$   & 0.0575     \\[0.5ex]
\hline 
\end{tabular}
\label{Params}
\end{table}

In terms of computation time, PAES was a bit faster than the other MOEAs, as it does not require the additional ranking mechanisms like non-dominated sorting and crowding distance comparison. Considering all the results, we can conclude that the proposed solution achieves results that are comparable to or even exceed those of existing multi-objective algorithms.

\section{Conclusions and Future Work}\label{section.conclusion}

Traditional feature selection methods often discard important features. To address this, we have employed two global search strategies: a decomposition-based binary GA in our existing work \cite{Ghahramani2020Smart1}, and a multi-objective evolutionary algorithm in this study, both aiming at identifying an optimal subset of features to improve classification accuracy. By integrating multi-objective optimization with a DNN, we offer a flexible, AI-driven solution tailored to smart manufacturing needs.

The decomposition model, while effective, treats the multi-objective problem as single-objective subproblems, which can be computationally expensive. To improve efficiency, we employed a meta-modeling approach, considering all objectives simultaneously. The multi-objective optimization algorithm approximated Pareto-optimal solutions with a tailored non-dominated sorting method. Our approach, which maintained diversity and ensured an even distribution of feature sets, outperformed previous solutions, enabling the exploration of all possible feature subsets in a single run. Note that, in this work, we consider two objectives; however, as the number of objectives increases, the search space becomes more complex and the proportion of non-dominated solutions rises, making it more difficult to distinguish truly optimal solutions.

From an implementation perspective, our model is well-suited for integration into smart manufacturing pipelines using modular AI frameworks. The use of evolutionary algorithms and batch processing provides an efficient foundation for handling larger datasets. Evolutionary algorithms are inherently suitable for exploring large solution spaces, while batch processing enables the model to manage large volumes of data without overloading system resources. This makes the approach adaptable to diverse industrial data, including high-dimensional and large-scale datasets. The computational resources required for the proposed evolutionary algorithm and neural network training are modest and can be scaled depending on dataset size. Large-scale deployment can be challenging, and parallel processing frameworks, such as Apache Spark or TensorFlow, are recommended for efficient data handling and real-time inference. Integrating the AI-driven model with existing legacy systems also poses challenges, particularly in ensuring data consistency from sensors across various production stages. It's crucial to address interoperability issues between new and old systems and adjust workflows to integrate AI-based decision-making effectively. Moreover, aligning the optimization framework with production scheduling systems and overcoming data heterogeneity are critical factors for successful deployment in complex manufacturing environments. Our future work aims to address the challenges related to data heterogeneity and interoperability. We also plan to explore strategies to mitigate the effects of imbalanced datasets \cite{Wang2023-Minority,Yan-BO-SMOTE}.



\ifCLASSOPTIONcaptionsoff
  \newpage
\fi

\begin{IEEEbiography}[{\includegraphics[width=1in,height=1.25in,clip,keepaspectratio]{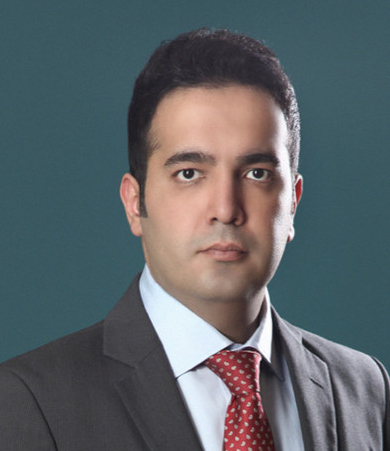}}]{Mohammadhossein Ghahramani}
(Senior Member) obtained his B.S. and M.S. degrees in Information Technology Engineering from Amirkabir University of Technology-Tehran Polytechnic. He earned his Ph.D. in Computer Technology and Application from Macau University of Science and Technology in 2018. He was a member of the Insight Centre for Data Analytics and a Research Fellow at University College Dublin, Ireland. Currently, he is an Assistant Professor of Data Science at Birmingham City University, UK. His research interests include smart systems, artificial intelligence, optimization, smart cities, and IoT. Dr Ghahramani has published various peer-reviewed journal papers in reputable journals and has received several awards, including the Best Automation Paper in Technology by the IEEE Robotics and Automation Society. He serves as a co-chair of the IEEE SMCS Technical Committee on AI-based Smart Manufacturing Systems and as an Associate Editor of IEEE Internet of Things Journal.\end{IEEEbiography}
\vfil\vfil\vfil\vfil\vfil
\vfil

\begin{IEEEbiography}[{\includegraphics[width=1in,height=1.25in,clip,keepaspectratio]{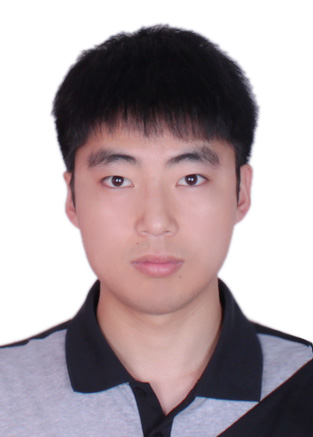}}]{Yan Qiao}(Senior Member)
received the B.S. and Ph.D. degrees in industrial engineering and mechanical engineering from Guangdong University of Technology, Guangzhou, China, in 2009 and 2015, respectively. He is currently an Associate Professor with the Institute of Systems Engineering and the Department of Engineering Science, Faculty of Innovation Engineering, Macau University of Science and Technology. He has over 100 publications, including one book chapter and more than 50 regular articles in IEEE Transactions. Besides, he was a recipient of several awards. He is an Associate Editor of IEEE Robotics and Automation Magazine..
\end{IEEEbiography}

\begin{IEEEbiography}[{\includegraphics[width=1in,height=1.25in,clip,keepaspectratio]{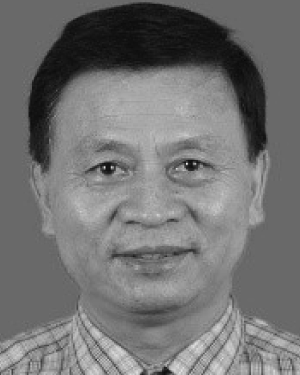}}]{NaiQi Wu}
(M'04-SM'05-F'18) received his B.S. degree from the Anhui University of Technology, in 1982, M.S. and Ph.D. degrees from Xi’an Jiaotong University, in 1985 and 1988, respectively. He joined the Shenyang Institute of Automation, Chinese Academy of Sciences, Shenyang, China and became a faculty member of Shantou University, Shantou, China in 1995. He became a professor of the Guangdong University of Technology, Guangzhou, China, in 1998. He founded the Institute of Systems Engineering at Macau University of Science and Technology, Taipa, Macau, in 2013 and currently is a Chaired Professor. He has published one book, five book chapters, and over 250 peer-reviewed journal papers. He was an Editor-in-Chief of different reputable journals.
\end{IEEEbiography}

\begin{IEEEbiography}[{\includegraphics[width=1in,height=1.25in,clip,keepaspectratio]{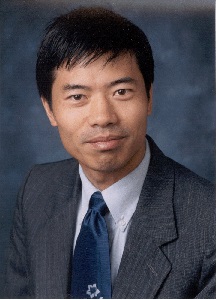}}]{MengChu Zhou}
(S'88-M'90-SM'93-F'03)(Fellow, IEEE) received his Ph. D. degree from Rensselaer Polytechnic Institute, Troy, NY in 1990 and then joined New Jersey Institute of Technology where he has been Distinguished Professor since 2013. His interests are in Petri nets, automation, robotics, big data, Internet of Things, cloud/edge computing, and AI.  He has 1400+ publications including 17 books, 900+ journal papers (700+ in IEEE transactions), 31 patents and 32 book-chapters. He is Fellow of IFAC, AAAS, CAA and NAI.
\end{IEEEbiography}

\vfill


\end{document}